%% file: main.tex
\documentclass[10pt,twocolumn,letterpaper]{article}

\usepackage{cvpr}              
\usepackage{adjustbox}
\usepackage{subcaption}
\usepackage{amsfonts}
\usepackage{amsmath}
\usepackage{amssymb}
\usepackage{ragged2e} 
\usepackage{booktabs,makecell, multirow, tabularx}
\usepackage{float}
\usepackage{graphicx} 
\usepackage[table]{xcolor}
\usepackage{pifont}
\usepackage{epstopdf} 
\usepackage{bibentry}
\usepackage{xurl}
\input{preamble}
\definecolor{cvprblue}{rgb}{0.21,0.49,0.74}
\usepackage[pagebackref,breaklinks,colorlinks,allcolors=cvprblue]{hyperref}

\def\paperID{} 
\def\confName{CVPR}
\def\confYear{2026}

\title{Rel-Zero: Harnessing Patch-Pair Invariance for Robust Zero-Watermarking Against AI Editing}

\def\authorBlock{
Pengzhen Chen\textsuperscript{1,2,3} \quad 
Yanwei Liu\textsuperscript{1,3}\textsuperscript{\textdagger} \quad 
Xiaoyan Gu\textsuperscript{1,2,3}\textsuperscript{\textdagger} \quad Xiaojun Chen\textsuperscript{1,2,3} \quad 
Wu Liu\textsuperscript{4} \quad 
Weiping Wang\textsuperscript{1}\\

    \textsuperscript{1}Institute of Information Engineering, Chinese Academy of Sciences \\
    \textsuperscript{2}School of Cyber Security, University of Chinese Academy of Sciences \\
    \textsuperscript{3}State Key Laboratory of Cyberspace Security Defense  \quad
    \textsuperscript{4}University of Science and Technology of China\\ 
    {\tt\small \{chenpengzhen, liuyanwei, guxiaoyan, chenxiaojun, wangweiping\}@iie.ac.cn, liuwu@ustc.edu.cn}
}

\begin{document}
\twocolumn[{

\renewcommand\twocolumn[1][]{#1}
\maketitle
\begin{center}
    \vspace{-4\baselineskip}
    \author{\authorBlock}
\end{center}
}]

\input{sec/0_abstract}

{
  \renewcommand{\thefootnote}%
    {\fnsymbol{footnote}}
  \footnotetext[0]{ $\dagger$ Corresponding author. 
  }
  }
\vspace{-1em}

\input{sec/1_intro}
\input{sec/2_related}

\input{sec/3_theory}
\input{sec/4_method}

\input{sec/5_experiments}

\input{sec/6_conclusion}
\input{sec/acknowledgement}
{
    \small
    \bibliographystyle{ieeenat_fullname}
    \bibliography{main}
}

\input{sec/X_suppl}

\end{document}

%% file: sec/0_abstract.tex
\begin{abstract}

Recent advancements in diffusion-based image editing pose a significant threat to the authenticity of digital visual content. Traditional embedding-based watermarking methods often introduce perceptible perturbations to maintain robustness, inevitably compromising visual fidelity. Meanwhile, existing zero-watermarking approaches, typically relying on global image features, struggle to withstand sophisticated manipulations. In this work, we uncover a key observation: while individual image patches undergo substantial alterations during AI-based editing, the relational distance between patch pairs remains relatively invariant. Leveraging this property, we propose Relational Zero-Watermarking (Rel-Zero), a novel framework that requires no modification to the original image but derives a unique zero-watermark from these editing-invariant patch relations. By grounding the watermark in intrinsic structural consistency rather than absolute appearance, Rel-Zero provides a non-invasive yet resilient mechanism for content authentication. Extensive experiments demonstrate that Rel-Zero achieves substantially improved robustness across diverse editing models and manipulations compared to prior zero-watermarking approaches. Code is available at \url{https://github.com/JerCCC7/Rel-Zero}.

\end{abstract}

%% file: sec/1_intro.tex
\section{Introduction}
\label{sec:intro}

The rapid development of sophisticated generative models, particularly diffusion models~\cite{ldm, ldm1,ldm2,ldm3}, has revolutionized digital content creation. Editing frameworks such as InstructPix2Pix~\cite{edit1} and inpainting systems \cite{ctrln,edit2} empower users to edit images with unprecedented ease and realism. However, they also raise exigent challenges in content authenticity, copyright protection, and provenance tracking. Digital watermarking serves as a critical technology to address these concerns, yet existing paradigms, embedding watermarking and zero-watermarking, struggle to balance robustness, fidelity, and applicability in the face of generative edits.

\begin{figure}[t]
  \centering
  \includegraphics[width=\linewidth]{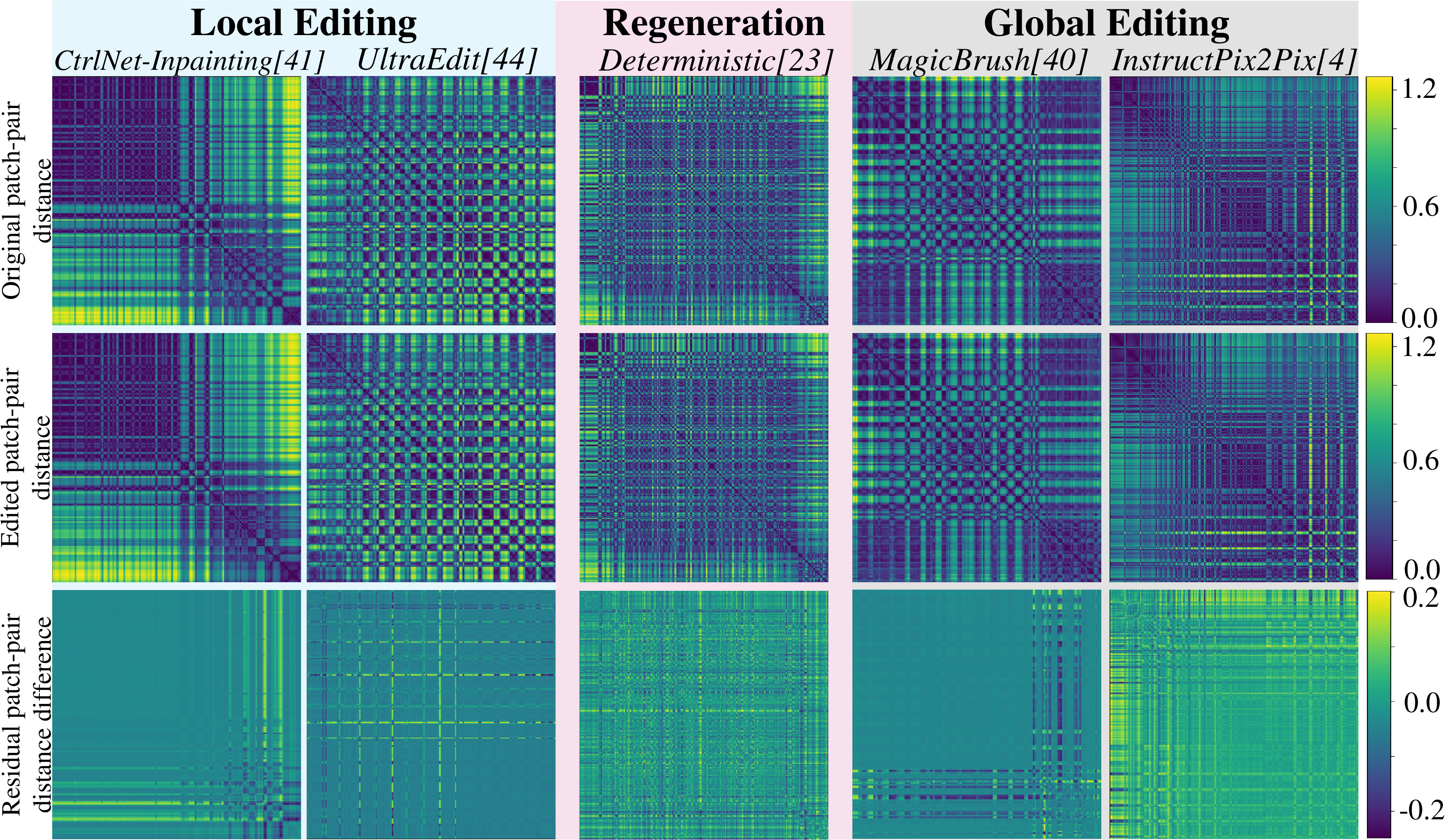}
  \caption{\textbf{Analysis of Relational Stability.} We reveal a key insight: Patch-pair distance tends to be preserved after AI editing. This invariant property can be extracted as a zero-watermark. Here we present the patch-pair distance of RGB vectors in original images (first row) and edited images (second row) across five main editing models. The last row illustrates the distance difference between pre-edit and post-edit with a color scale spanning [-0.2,0.2].
  }
  \label{fig:INTRO}
\end{figure}

Traditional embedding watermarking paradigms~\cite{deep1,deep3,editguard,cin} inject watermark signals into the image pixels or frequency domain. To survive aggressive manipulations, especially non-linear generative edits that fundamentally reconstruct pixel data, the embedded signal must be sufficiently strong~\cite{robustwide,vine,trustmark}. This requirement inevitably sacrifices image quality, leading to perceptible artifacts and degradation of the original content. Across domains such as medical imaging~\cite{medical1,medical2,medical3,medical4}, autonomous driving~\cite{ad1,ad2}, and world modeling~\cite{world1,world2}, copyright protection is paramount, yet compromised image fidelity can be catastrophic. The introduction of watermark-induced noise can severely impair the reliability of downstream tasks, which leads to inaccurate diagnosis, deviations in pathological analysis, unstable perception, and distorted environment prediction, ultimately limiting the trustworthiness of high-level visual understanding systems. Therefore, this trade-off is intolerable for these creative workflows. Given the high copyright value of images in these fields, protection that ensures robustness against edits without compromising image quality is essential.

On the other hand, zero-watermarking methods~\cite{zero1,zero2,zero4,plugmark} offer a compelling alternative by preserving pristine image quality. These methods extract a unique fingerprint or key from the original image and store it externally without altering the image structure. Verification involves checking if a suspect image still possesses this key. While elegant, the robustness of current zero-watermarks is critically low against generative edits. Existing methods~\cite{oldzero1,zero3,oldzero2} primarily rely on fragile high-frequency details, global statistics, or absolute feature descriptors (e.g., from SIFT or deep classifiers), which are precisely what generative models are designed to alter or completely restructure, rendering the watermarks invalid.
To address these challenges, we conduct an empirical analysis on images before and after editing. Our findings reveal that, while absolute features and pixel values of an image are drastically altered by generative edits, the \textbf{intrinsic relational structure} between partial image regions remains remarkably stable. Specifically, as shown in Fig.~\ref{fig:INTRO}, we discover that the relative distance between certain pairs of image patches, measured in RGB space, is largely invariant even after complex edits like style transfer or instruction-based manipulation, i.e., if patch $i$ and patch $j$ are ``very different" (large RGB distance) in the original image, they tend to remain ``very different" after the edit.

Based on this discovery, we propose a novel zero-watermarking framework built upon \textbf{Stable Relational Geometry} to extract these invariant patch pairs. Instead of relying on fragile absolute features, our method identifies a unique and resilient set of patch pairs as the zero-watermark, whose relational distance is robustly preserved after substantial editing.
Our contributions are as follows:
\begin{itemize}
\item We identify and validate a key property for generative editing: the stability of patch-wise relational distance in feature space.
\item We propose a novel zero-watermarking framework to learn and predict this stable relational structure from a single image, formulating the watermark as a set of robust patch pairs.
\item Experiments demonstrate that our zero-watermarking approach achieves promising robustness against a wide range of generative edits, while inherently maintaining perfect image fidelity.
\end{itemize}

%% file: sec/2_related.tex
\section{Related Work}
\label{sec:related}

\subsection{Conventional Embedding Watermarking}

Traditional watermarking techniques~\cite{dwt} embed a signal directly into the pixel domain or frequency domain (e.g., DCT, DWT). While effective for simple copyright marking, these methods are fragile and can be easily broken by geometric transformations, compression, and especially the non-linear reconstruction performed by generative models.

More recently, deep learning-based embedding methods~\cite{deep1,deep3,drgw,trustmark,sd2} have shown improved performance. They train neural networks to encode messages into images through deep features, and have demonstrated strong robustness against a wide range of distortions(e.g., compression, scaling and noise). In the era of AIGC, to survive diffusion-based AI-editing, several state-of-the-art methods, such as VINE~\cite{vine} and RobustWide~\cite{robustwide}, are proposed specifically to withstand these edits. Their strategies involve injecting a powerful, resilient signal into perceptually significant, low-frequency components of the image. However, this introduces an unavoidable trade-off. To achieve robustness against generative models that fundamentally alter pixel data, the embedded signal must be strong, which \textbf{inevitably degrades image fidelity}. As a result, such methods are inappropriate for scenarios where preserving pristine visual quality is essential.

\subsection{Zero-Watermarking}

Zero-watermarking~\cite{oldzero1} is designed to fundamentally address the fidelity issue. Instead of embedding any signal into the image, it extracts a unique ``fingerprint" from the original image and stores it externally (e.g., in a database) alongside the content's metadata. Verification is then performed by extracting a fingerprint from a suspect image and comparing it to the stored reference. Early methods~\cite{oldzero2} rely on fragile handcrafted features such as SIFT or SURF. Recent approaches~\cite{zero1,zero2,zero3,zero4} leverage robust deep features from pre-trained networks to achieve improved robustness.

Despite guaranteeing perfect fidelity by design, zero-watermarking exhibits \textbf{critically limited robustness} facing modern generative edits. Unlike traditional perturbations, generative models do not simply inject noise. They reconstruct regions, rewrite textures, and modify semantic content. Such transformations fundamentally alter or destroy the absolute feature descriptors and spatial structures that current zero-watermarks rely on. Consequently, existing zero-watermarking techniques fail catastrophically under these modern edits. This exposes a significant gap in the field: the absence of a watermarking paradigm that simultaneously preserves the perfect fidelity of zero-watermarking and delivers the robustness required to withstand generative image manipulation. Our work is designed precisely for it.

%% file: sec/3_theory.tex
\begin{figure*}[htbp]
  \centering
  \subfloat[The distribution of residual $ |d^{\text{after}}_{ij} - d^{\text{before}}_{ij}|$. The majority is tightly distributed around zero.]{
    \includegraphics[width=\linewidth]{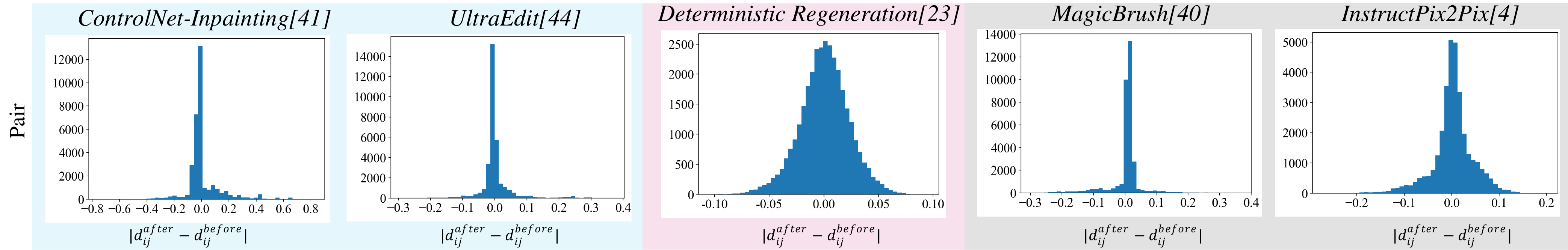}
    \label{fig:ssm_A}
  }\hfill
  \subfloat[\textbf{Pairwise RGB distance preservation.}
    Distance after editing ($d^{\text{after}}$) vs. before editing ($d^{\text{before}}$) for all RGB patch pairs. The data is fitted with a global linear model $y = \alpha x + \beta$.
    We report the fitted $\hat{\alpha} \approx 1$, a high coefficient of determination (e.g., $R^2 > 0.95$), and a high Spearman rank correlation ($\rho > 0.98$).]{
    \includegraphics[width=\linewidth]{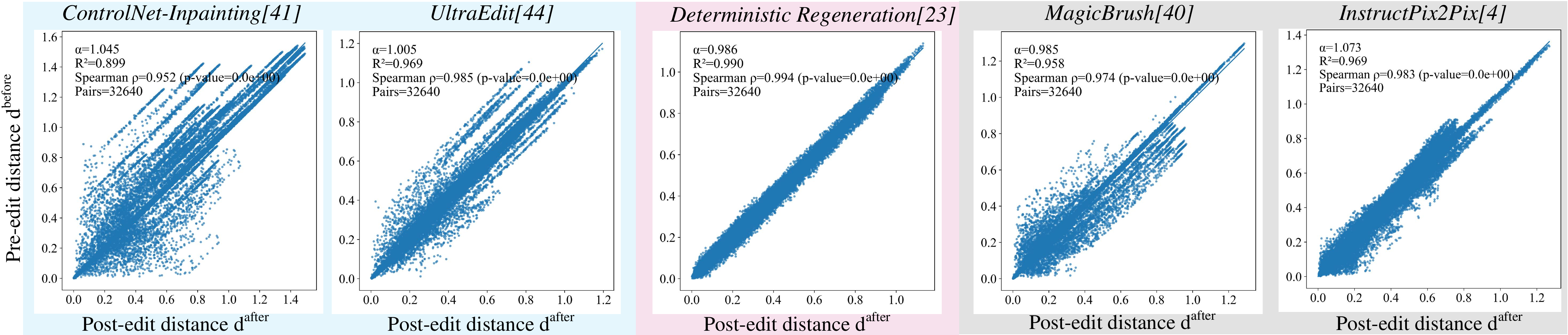}
    \label{fig:ssm_B}
  }
  \caption{\textbf{Analysis of Relational Stability.}}
  \label{fig:ssm}
\end{figure*}
\section{Pairwise Patch Distance Preservation Under Editing}
\subsection{Experimental Discovery}
\label{sec:theory}
We present how the invariance relationship is discovered between patch pairs before and after AI editing. Our experiments demonstrate that, after AI editing, the distance between corresponding patch pairs remains stable and exhibits a relatively invariant pattern.

\textbf{Evaluation data:}
We randomly sample 10,000 instances from UltraEdit~\cite{ultraedit} and MagicBrush~\cite{edit3} datasets. Of these samples, 2,000 are allocated for deterministic regeneration, another 4,000 for global editing, and 4,000 for local editing.

\textbf{Pairwise Patch Modeling:}
We model the images as a set of non-overlapping $N = 256$ patches, represented by their mean RGB vectors $\{v_i\}_{i=1}^{N}$. The ``internal color structure" can be characterized by the $\binom{N}{2}$ pairwise distance in this feature space.
Let $d^{\text{before}}_{ij} = \|v_i - v_j\|$ be the L2-distance between the mean RGB vectors of patches $i$ and $j$ in the original image, and $d^{\text{after}}_{ij} = \|v'_i - v'_j\|$ be the distance after editing. Consequently, the difference between the distance before and after
editing is defined as $|d^{\text{before}}_{ij} - d^{\text{after}}_{ij}|$.

\textbf{Key Observation:} \textit{The \textit{relative distance} between particular pairs of patches appears remarkably invariant, even if the edit operation (transformation $\mathcal{T}$) substantially alters the RGB values of individual patches (mapping $v_i$ to $v'_i$).}

\textbf{Experimental Validation:} We conduct experiments to analyze distance of RGB vectors before and after three mainstream AI editing methods, including local editing (ControlNet-Inpainting\cite{ctrln}, UltraEdit\cite{ultraedit}), global editing(InstructPix2Pix\cite{edit1}, MagicBrush\cite{edit3}) and deterministic regeneration~\cite{det}.
Fig.~\ref{fig:INTRO} depicts the overall variation in patch-pair RGB vectors distance before and after editing and their differences. 
Furthermore, Fig.~\ref{fig:ssm_A} illustrates the numerical distribution of these differences. 

We observe that the differences exhibit a near–zero-mean, tightly distributed pattern with no evident systematic bias.
Besides, the majority of patch-pairs exhibit distance differences clustered closely around 0.0. These results provide strong evidence that a substantial portion of patch-pair distance remains effectively invariant across the editing process, indicating that intra-image geometric consistency is largely preserved.

To rigorously quantify this phenomenon, we move beyond anecdotal observation and propose a formal geometric hypothesis. 

\textbf{Hypothesis:} If the editing transformation $\mathcal{T}$ is indeed ``structure-preserving," it should act as a predictable, geometrically simple operation on this set of pairwise distance. The strongest form is that $\mathcal{T}$ induces a \textbf{global similarity transformation} on the feature-space manifold. Therefore, we propose a simple and direct hypothesis:
\textit{\textbf{does a global linear relationship 
exist between the pre-edit and post-edit distance?}}

To validate it, we conduct a large-scale distance-distance correlation analysis. For a given edited image, we compute all $\binom{N}{2}$ pairwise distance $d^{\text{before}}_{ij}$ and $d^{\text{after}}_{ij}$. We then visualize their relationship by plotting $d^{\text{after}}_{ij}$ (y-axis) against $d^{\text{before}}_{ij}$ (x-axis) and perform a linear regression:
\begin{equation}
    d^{\text{after}}_{ij} \approx \alpha \cdot d^{\text{before}}_{ij} + \beta,
    \label{eq:linear_model}
\end{equation}
where $\alpha$ denotes a global scaling factor (a ``stretching" or ``shrinking" of the feature space), and $\beta$ represents a uniform offset.

Fig.~\ref{fig:ssm_B} provides compelling support for our hypothesis. The fitted slope $\alpha \approx 1$, offset $\beta \approx 0$, together with the high coefficient of determination ($R^2 > 0.95$) and near-perfect Spearman correlation ($\rho \approx 1$) indicate an exceptionally tight linear correlation between $d^{\text{after}}_{ij}$ and $d^{\text{before}}_{ij}$. It demonstrates that most of the variance in post-edit distance can be directly explained by their pre-edit counterparts. More broadly,
this phenomenon reveals a near–\textbf{affine invariance} in the feature space: the relative distance among partial patches is preserved up to a uniform scale.

\subsection{Theoretical Justification}
We delve deeper into our observations to explore why this stability exists. This phenomenon can be attributed to two possible properties of generative editing models:

\begin{enumerate}
    \item \textbf{Fidelity Constraints and Regularization:} Generative editing models, especially diffusion models, are not trained in an unconstrained manner. They explicitly or implicitly incorporate ``content/structure preservation" losses. These include perceptual losses (LPIPS)~\cite{lpips}, reconstruction losses ($L_1/L_2$ in latent or pixel space), and various consistency regularizers. Such optimization objectives strongly penalize unnecessary perturbations and encourage the model to learn an identity mapping for non-target regions. As a result, the cross-patch relationships (governing shape, layout, and texture ratios) become a core invariant that these models are optimized to preserve. 

    \item \textbf{Low-Dimensionality of the Edit Subspace:} Semantic edits (e.g., ``change style", ``make it summer") typically correspond to low-dimensional directions or sub-manifolds within the model's latent space $z$ ~\cite{hertz2022prompt}. When such a latent transformation $\mathcal{T}$ (e.g., $z' = \mathcal{T}(z)$) is decoded, it often applies a functionally uniform change across the image statistics. For example, a style change may apply a similar color-space shift or texture basis change to all patches. When this transformation is approximately affine in the feature space ($v'_i \approx A v_i + b$), the difference between two patch features becomes $v'_i - v'_j \approx A(v_i - v_j)$. The magnitude of this difference is scaled ($||v'_i - v'_j|| \approx \alpha ||v_i - v_j||$), but the fundamental relationship is preserved and naturally leads to the strong linear predictability in our empirical analysis.
\end{enumerate}

Together, these findings suggest that patch-pair relational stability is an intrinsic signature during semantic editing, a property we can exploit for robust zero-watermarking. Since this relational geometry is stable and scalable invariant, if we can reliably identify these stable pairs, we can construct a zero-watermark based on these pair indices. 

%% file: sec/4_method.tex
\begin{figure*}[t]
  \centering
  \includegraphics[width=\linewidth]{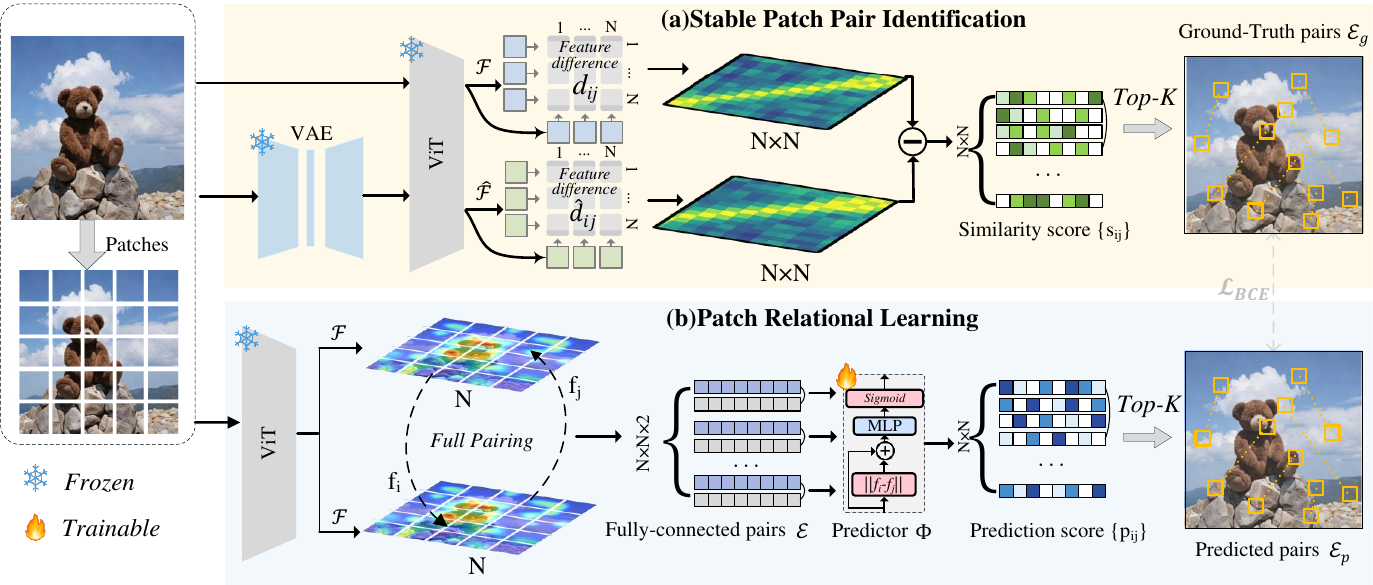}
  \caption{\textbf{Framework of the proposed Rel-Zero.}
  \emph{(a) Stable Patch Pair Identification.} To train a predictor capable of identifying patch pairs with invariant distance relationships, we first construct training targets.
A pretrained VAE is employed to simulate generative edits.
Features of the original and VAE-modified images are extracted using a ViT to obtain patch-wise features $\mathcal{F}$ and $\mathcal{\hat F}$.
Pairwise distance is computed on both feature maps, and their differences are measured to identify the most stable top-$K$ pairs surviving  edits, which serve as the ground-truth pairs $\mathcal{E}_g$.
(b) \emph{Patch Relational Learning.}
Given an input image, ViT features are extracted and all patch pairs are densely formed to construct fully-connected pairs $\mathcal{E}$.
A learnable pair predictor $\Phi$ then estimates the stability scores of each pair, which are trained to align with the ground-truth pairs $\mathcal{E}_g$ from  (a).
During inference (i.e., watermark generation and verification), only the module in (b) is required to generate relational zero-watermarks $\mathcal{E}_p$.
}
  \label{fig:framework}
\end{figure*}

\section{Methodology}
\label{sec:method}

In this section, we introduce the proposed \textbf{Relational Zero-Watermarking (Rel-Zero)} framework, a robust zero-watermarking approach that leverages relational invariance among patch-pair distance to resist generative manipulations. As shown in Fig~\ref{fig:framework}, the framework is composed of three key stages:
(1) \textit{Stable patch-pair identification} to identify the ground-truth invariant pairs as zero-watermark targets;
(2) \textit{Patch Relational Learning} to build a network to predict these invariant pairs; and
(3) \textit{Watermark Generation and Verification}. 

\subsection{Stable Patch Pair Identification}
\label{ssec:ground_truth}

We first require a set of ``ground-truth" stable patch pairs as the training target. Inspired by \cite{editguard}, we observe that the structural impact of diffusion-based generative edits on patch-pair relationships is analogous to that of its Variational Autoencoder (VAE) component, while the latter is substantially more computationally friendly. Therefore, we employ a pre-trained VAE~\cite{vae} to simulate the essential content-preserving and smoothing properties of generative pipelines.

The goal is to identify patch pairs $(i, j)$ whose feature-space relationships remain stable before and after VAE reconstruction. Given an input image $\mathbf{I}$, we first pass it through the VAE to obtain a reconstructed version $\hat{\mathbf{I}} =V_{ae}(\mathbf{I})$.

Both the original image $\mathbf{I}$ and the reconstructed $\hat{\mathbf{I}}$ are partitioned into $N$ non-overlapping patches. Each patch set is then fed into a pre-trained Vision Transformer~\cite{vit} (ViT)  $\phi_{\text{vit}}$ to extract high-dimensional patch-level representations:
\begin{equation}
\mathcal{F} = \phi_{\text{vit}}(\mathbf{I}), \quad \hat{\mathcal{F}} = \phi_{\text{vit}}(\hat{\mathbf{I}}),
\end{equation}
where $\mathcal{F} = [\mathbf{f}_1, \ldots, \mathbf{f}_\mathbf{N}]$ and $\hat{\mathcal{F}} = [\hat{\mathbf{f}}_1, \ldots, \hat{\mathbf{f}}_\mathbf{N}]$ denote the patch-level embeddings. (Further ViT feature analysis in Appendix Sec.~\ref{sec:vitanalysis})

For each patch pair $(i,j)$, we define their feature difference (L2 distance) before and after reconstruction as:
\begin{equation}
d_{ij} = \|\mathbf{f}_i - \mathbf{f}_j\|_2,
\quad \hat{d}_{ij} = \|\hat{\mathbf{f}}_i - \hat{\mathbf{f}}_j\|_2.
\end{equation}
The stability of a relation is then measured by a similarity score: 
\begin{equation}
    s_{ij} = \exp\left( - |d_{ij}  - \hat{d}_{ij}| \right).
\end{equation}

The top-$K$ pairs $(i, j)$ with the highest similarity scores $s_{ij}$ are chosen as the ground-truth stable set. These pairs constitute the \textit{invariant relations} $\mathcal{E}_{\text{g}}$, which serve as the ground-truth target for our zero-watermark.

\subsection{Patch Relational Learning}
\label{ssec:graph_selector}

To learn the relational structure from the image features, 
we construct a fully connected set of patch pairs $\mathcal{E}$ from the ViT-extracted patch features. 
Each pair $(i,j )\in \mathcal{E}$ corresponds to the relation between patch $i$ and $j$, where $i, j \in \{1,...,N\}$.

We employ a simple edge predictor $\Phi(\cdot)$ to estimate a prediction score $p_{ij}$ for each pair:
\begin{equation}
p_{ij} =  \Phi(\mathbf{f}_i, \mathbf{f}_j)= \sigma\big( \psi(\mathbf{f}_i \oplus \mathbf{f}_j \oplus \, \|\mathbf{f}_i - \mathbf{f}_j\|_2) \big),
\end{equation}
where $\sigma(\cdot)$ denotes the sigmoid activation, $\psi(\cdot)$ represents a MLP network, and $\oplus$ indicates concatenation.




\subsection{Training Objective}

The training target aims to align the learned relational scores $\{p_{ij}\}$ with the previously selected invariant pairs. We assign binary labels $y_{ij} = 1$ for the top-$K$ invariant edges and $y_{ij} = 0$ otherwise. The model is trained using the binary cross-entropy (BCE) loss:
\begin{equation}
    \mathcal{L}_{BCE} = - \frac{\sum_{i \neq j} \left[ y_{ij} \log(\hat{y}_{ij}) + (1 - y_{ij}) \log(1 - \hat{y}_{ij}) \right]}{N(N-1)} .
\end{equation}

\subsection{Watermark Generation and Verification}

\label{ssec:generation_verification}
\paragraph{Generation.}Given an image $\mathbf{I}$, the relational zero-watermark is defined as the top-$K$ most confident pairs predicted by $\Phi$:
\begin{equation}
\mathcal{E}_p = \text{Top-K}\big(\Phi(\phi_{\text{vit}}(\mathbf{I})) \big),
\end{equation}
where $\mathcal{E}_p = \{(i_1, j_1), \ldots, (i_K, j_K)\}$.
This index set can be securely stored externally (e.g., hashed or encrypted in a database (detailed in Appendix Sec.~\ref{sec:encrypt})) without revealing the original content.

\paragraph{Verification.}To verify a suspect image $\mathbf{I'}$, we extract its corresponding pairs:
\begin{equation}
\mathcal{E}_p' = \text{Top-K}\big( \Phi(\phi_{\text{vit}}(\mathbf{I'})) \big),
\end{equation}

and evaluate the similarity score $\eta$ between $\mathcal{E}_p$ and $\mathcal{E}_p'$:
\begin{equation}
\eta = \frac{|\mathcal{E}_p \cap \mathcal{E}_p'|}{K},
\label{eq:metric}
\end{equation}
where $|\cdot|$ denotes set cardinality (i.e., Jaccard-style overlap ratio of edge indices).
The image is deemed authenticated at an operating point calibrated for a target false-alarm rate 
(detailed in Appendix Sec.~\ref{sec:tpr}). 

%% file: sec/5_experiments.tex
\begin{figure*}[t]
  \centering
  \includegraphics[width=\linewidth]{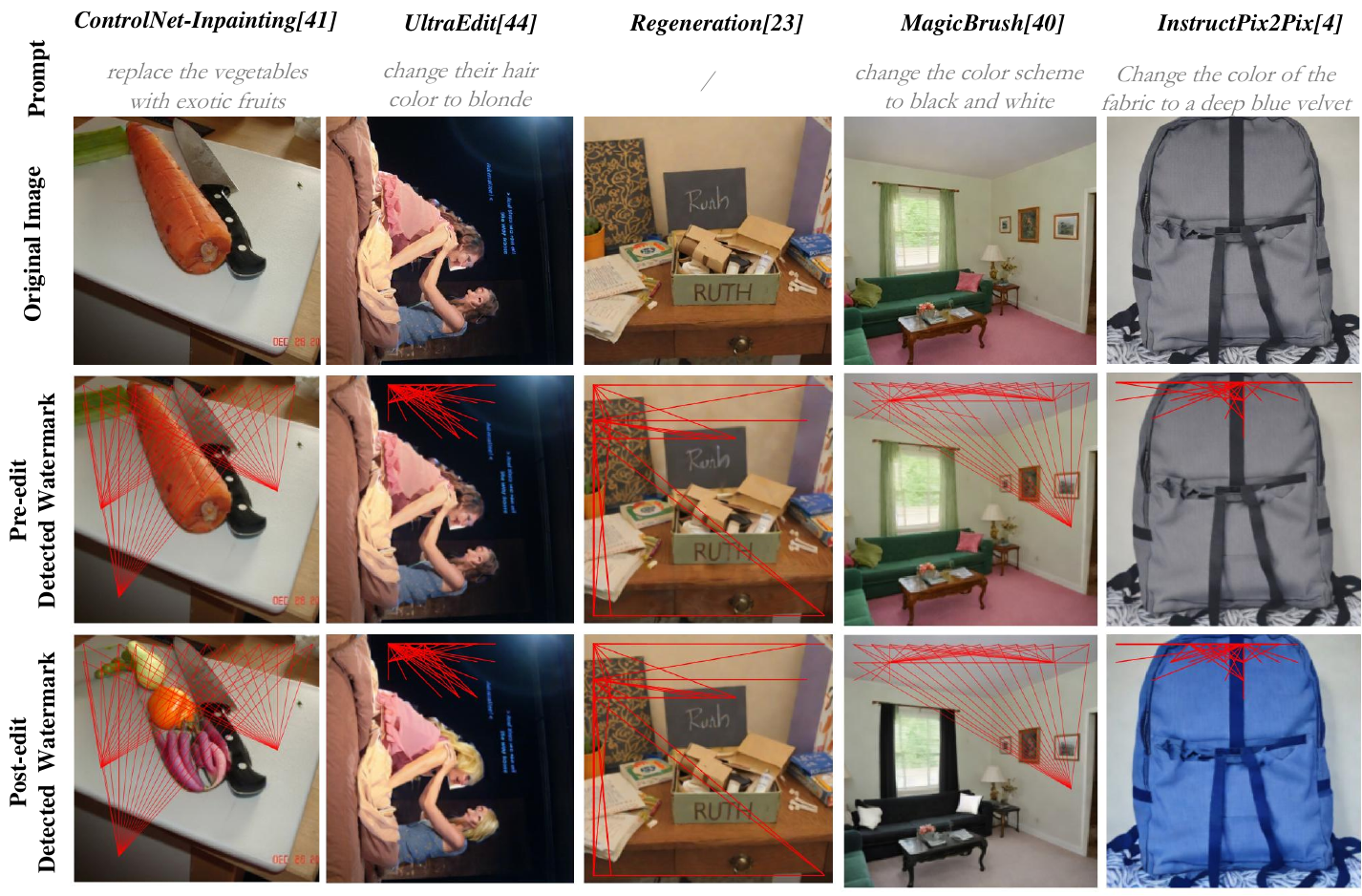}
  \caption{
  Visualization of pre-edit and post-edit extracted watermarks.}
  \label{fig:visual}
\end{figure*}
\section{Experiments}
\subsection{Experimental Setup}
\label{sec:setup}

\textbf{Datasets.}
We conduct our training and evaluation using diverse large-scale image datasets. The network is trained on the COCO\cite{coco} dataset.  For evaluation, we sample 10,000 instances from the widely-used UltraEdit~\cite{ultraedit} and MagicBrush~\cite{edit3} datasets as the validation
set. Each instance contains a source image, an editing prompt, and a region mask. All images are resized to
224 × 224 resolution. To ensure a fair and comprehensive
evaluation across a diverse range of image content, of these
images, 1,000 are allocated for the regeneration task, 1,000
for global editing, 4,000 for local editing, and 4,000 for common degradations.

\textbf{Baselines.}
For a thorough evaluation, we compare Rel-Zero to both embedding and zero-watermarking methods. Embedding watermarking includes DWT-DCT\cite{dwt} and the recent state-of-the-art VINE~\cite{vine} and RobustWide~\cite{robustwide}, which are designed against AI editing. For zero-watermarking, since only a few methods are open-sourced, we include ConZWNet~\cite{zero1} and FGPCET~\cite{zero2}, considering their timeliness and representativeness. All baselines are from their officially released checkpoints.

\paragraph{Attack Suite.}
We subject images to both Generative Editing and common distortions. The implementation details are in Appendix Sec.~\ref{sec:setting}. 
\begin{itemize}
\item Generative Edit Models.
To comprehensively evaluate robustness, we evaluate against a suite of modern generative editing models. We include InstructPix2Pix~\cite{edit1} and MagicBrush~\cite{edit3} for global editing, UltraEdit~\cite{ultraedit} and ControlNet-Inpainting~\cite{ctrln} for local editing. We also test against deterministic regeneration\cite{det}.
\item Common Distortions. Standard image processing operations including Random Cropout (50\% of the area), Scaling (to 0.5$\times$ and back), Contrast adjustments $\in[0.5,2.0]$, Brightness adjustments $\in[0.5,2.0]$, and Gaussian noise(std = 0.10).
\end{itemize}

\textbf{Evaluation Metrics.}
We evaluate all methods on two primary axes:
\begin{itemize}
    \item \textbf{Fidelity:} Although zero-watermarking does not affect image fidelity, we introduce PSNR\cite{psnr}, SSIM, and LPIPS\cite{lpips} to evaluate image quality for a fair assessment between embedding and zero-watermarking methods.
    \item \textbf{Robustness:}To assess watermark survivability, we report the True Positive Rate (TPR) at a highly stringent False{ Positive Rate (FPR) of 0.1\%.} This conservative threshold ensures high confidence in the uniqueness of watermark authentication. (detailed in Appendix Sec.~\ref{sec:tpr}.).
\end{itemize}

\textbf{Implementation Details.}
Unless specified otherwise, we use a patch size of $16 \times 16$, resulting in $N=196$ patches for a $224 \times 224$ image. We use ViT-B/16~\cite{vit} as our feature extractor $\phi_{\text{vit}}$ and utilize the VAE from Stable Diffusion v1.4\cite{ldm}. We set the watermark size to $K=50$ pairs. Experiments are conducted on an NVIDIA A100 GPU.

\begin{table*}[h]
\centering
\caption{
Comparison of visual fidelity (PSNR, SSIM, LPIPS) and robustness (TPR@(0.1\%FPR)) between Rel-Zero and baseline methods. 
Zero-watermarking methods do not modify image pixels, thus their PSNR metrics are marked as “–”.
Abbreviations: Det (Deterministic Regeneration), Pix2Pix (InstructPix2Pix), Ultra (UltraEdit), CtrlN (ControlNet-Inpainting).
}
\resizebox{\linewidth}{!}{%
\begin{tabular}{cccccccccccccc}
\toprule
\multirow{3}{*}{\textbf{Method}} &  
\multirow{3}{*}{\textbf{PSNR(dB)$\uparrow$}} &  
\multirow{3}{*}{\textbf{SSIM$\uparrow$}} &  
\multirow{3}{*}{\textbf{LPIPS$\downarrow$}} & 
\multicolumn{10}{c}{\textbf{TPR@(0.1\%FPR)(\%)$\uparrow$}} \\
\cmidrule(lr){5-14}
& & & &
\textbf{Regeneration} &
\multicolumn{2}{c}{\textbf{Global Editing}} &
\multicolumn{2}{c}{\textbf{Local Editing}} &
\multicolumn{5}{c}{\textbf{Common Distortions}} \\
\cmidrule(lr){5-5}
\cmidrule(lr){6-7}
\cmidrule(lr){8-9}
\cmidrule(lr){10-14}
& & & &
\textbf{Det} &
\textbf{Pix2Pix} & \textbf{Magic} &
\textbf{Ultra} & \textbf{CtrlN} &
\textbf{Cropout} & \textbf{Scaling} & \textbf{Contrast} & \textbf{Brightness} & \textbf{Gaussian}\\ 
\midrule

\multicolumn{14}{l}{\textbf{Embedding Watermarking}}\\
DWT-DCT \cite{dwt} 
& 40.38 & 0.9705 & 0.0136 
& 0.09 & 0.04 & 0.05 & 0.32 & 0.56 
& 10.35 & 6.78 & 30.18 & 51.88 & 12.45\\

Robust-Wide \cite{robustwide} 
& 41.93 & 0.9908 & 0.0034
& 90.41 & 97.23 & 81.97 & 80.45 & 82.11
& 95.31 & 96.45 & 98.93 & 98.89 & 98.12\\

VINE \cite{vine} 
& 37.34 & 0.9934 & 0.0063
& 99.98 & 97.46 & 94.58 & 99.96 & 93.04
& 54.87 & 76.43 & 98.43 & 97.90 & 98.37\\

\multicolumn{14}{l}{\textbf{Zero-Watermarking}}\\
ConZWNet \cite{zero1} 
& -- & 1.000 & 0.000
& 0.10 & 0.02 & 0.01 & 5.13 & 2.41
& 98.75 & 97.43 & 96.22 & 96.56 & 98.75\\

FGPCET \cite{zero2} 
& -- & 1.000 & 0.000
& 1.13 & 0.54 & 0.11 & 7.25 & 3.22
& 89.31 & 84.78 & 86.31 & 85.44 & 84.67\\

\rowcolor{gray!20}
\textbf{Rel-Zero} 
& -- & 1.000 & 0.000
& 85.13 & 89.65 & 95.63 & 96.55 & 97.43
& 98.45 & 98.57 & 96.45 & 97.93 & 95.12\\

\bottomrule
\end{tabular}
}
\label{tab:compare}
\end{table*}

\subsection{Evaluation Comparison}
\label{ssec:quantitative}

\textbf{Fidelity.}
As shown in Table~\ref{tab:compare}, the major advantage of zero-watermarking over embedding-based watermarking lies in its non-impact on visual quality. Although embedding watermarking methods can achieve reasonably high PSNR and SSIM values, they inevitably introduce slight distortions. In high-precision domains such as medical imaging, even minor distortions may lead to serious misinterpretations. Therefore, zero-watermarking provides a valuable solution for reliable copyright tracking of high-quality images. (Detailed comparisons in appendix Sec.~\ref{sec:quality}.)

\textbf{Robustness to Generative Edits.}
Fig.~\ref{fig:visual} illustrates the watermark extraction results before and after AI editing. 
Although these editing methods introduce significant structural and visual changes to the images, 
the difference relationships between certain patch pairs remain preserved, allowing successful watermark extraction. We observe that methods such as InstructPix2Pix and deterministic regeneration showcase larger deviations, this is consistent with our analysis in Fig.~\ref{fig:INTRO}, where these two methods perform large-scale global edits on images, which are more destructive compared to local editing operations.

Meanwhile, Table~\ref{tab:compare} reports the robustness of embedding-based and zero-watermarking methods under various AI editing operations. 
Among embedding approaches, Vine and Robust-Wide exhibit relatively strong performance, 
largely due to their adversarial training strategies that explicitly incorporate editing models during optimization. 
However, such strategies incur substantial computational costs, limiting their practicality in resource-efficient or general-purpose settings. In contrast, zero-watermarking baselines rely on invariant feature extraction. 
Yet these features are heavily distorted or fully overwritten by AI-editing, leading to the failure in watermark detection across diverse manipulation types.

\paragraph{Robustness to common distortions}
We further evaluate robustness against common image distortions. 
As shown in Table~\ref{tab:compare}, both zero-watermarking and embedding-based watermarking methods maintain strong performance under conventional perturbations. 
However, such robustness is typically achieved through extensive adversarial optimization, leading to increased training complexity. 
In contrast, Rel-Zero naturally maintains high resilience to distortions such as Gaussian noise, scaling, contrast, and brightness adjustments.
These operations introduce largely uniform transformations across the entire image, which may scale patch-pair distance but do not fundamentally alter their relative relationships. As discussed in Sec.~\ref{sec:method}, this preserves the patch-pair relational geometry, allowing Rel-Zero to maintain high pair-prediction accuracy and enabling stable zero-watermark extraction even under significant degradations. (Additional experiments in Appendix Sec.~\ref{sec:moreexperiment}.)

\begin{figure}[t]
\centering
\includegraphics[width=0.75\linewidth]{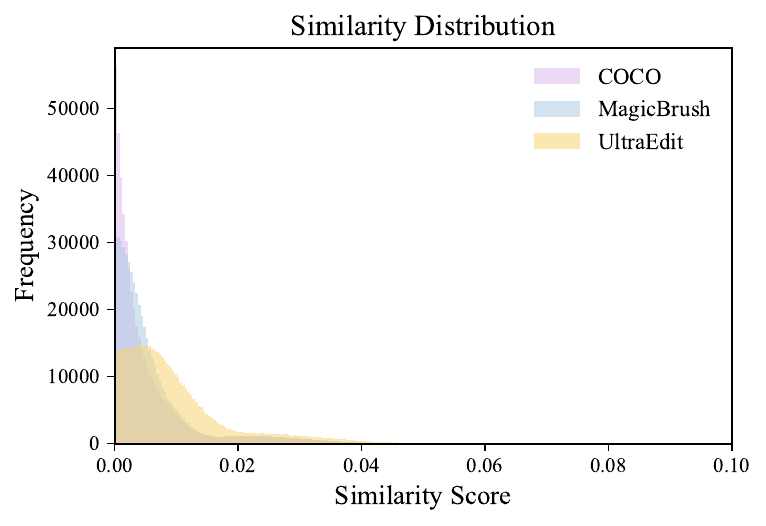}
\caption{Watermark similarity distribution of different images.}
\label{fig:uniq_cr}
\end{figure}

\subsection{Uniqueness \& Collision Analysis}
\label{sec:exp_uniqueness}
Beyond robustness and fidelity, another essential property of zero-watermarking is \textbf{uniqueness} across distinct images. Given two different images $\mathbf{I}_a \neq \mathbf{I}_b$ with their predicted top-$K$ edge sets $\mathcal{E}_p(\mathbf{I}_a)$ and $\mathcal{E}_p(\mathbf{I}_b)$ (Eq.~\ref{eq:metric}), we define their inter-image similarity as
\begin{equation}
\eta_{a,b}=\frac{|\mathcal{E}_p(\mathbf{I}_a)\cap \mathcal{E}_p(\mathbf{I}_b)|}{K}.
\end{equation}
Ideally $\eta_{a,b}\!\approx\!0$ for $\mathbf{I}_a\!\neq\!\mathbf{I}_b$.
We quantify it on three datasets---COCO, UltraEdit, and MagicBrush to cover diverse visual domains. 
For each dataset, we randomly sample 1{,}000 non-overlapping images, extract $\mathcal{E}_p$ with a fixed $K=50$, and compute the pairwise similarity $\eta_{a,b}$ among all image pairs. 
As illustrated in Fig.~\ref{fig:uniq_cr}, inter-image similarity remains consistently low across all datasets: 
most $\eta_{a,b}$ values concentrate near zero with minimal variance, and even the rare pairs showing slight correlation exhibit very small magnitudes, effectively ruling out any duplication. 
These results demonstrate that the learned relational pairs function as \emph{image-specific signatures} rather than generic, content-agnostic templates.

\begin{figure}[tbp]
  \centering
  \begin{subfigure}{0.5\textwidth}
    \centering
  \includegraphics[width=0.75\linewidth]{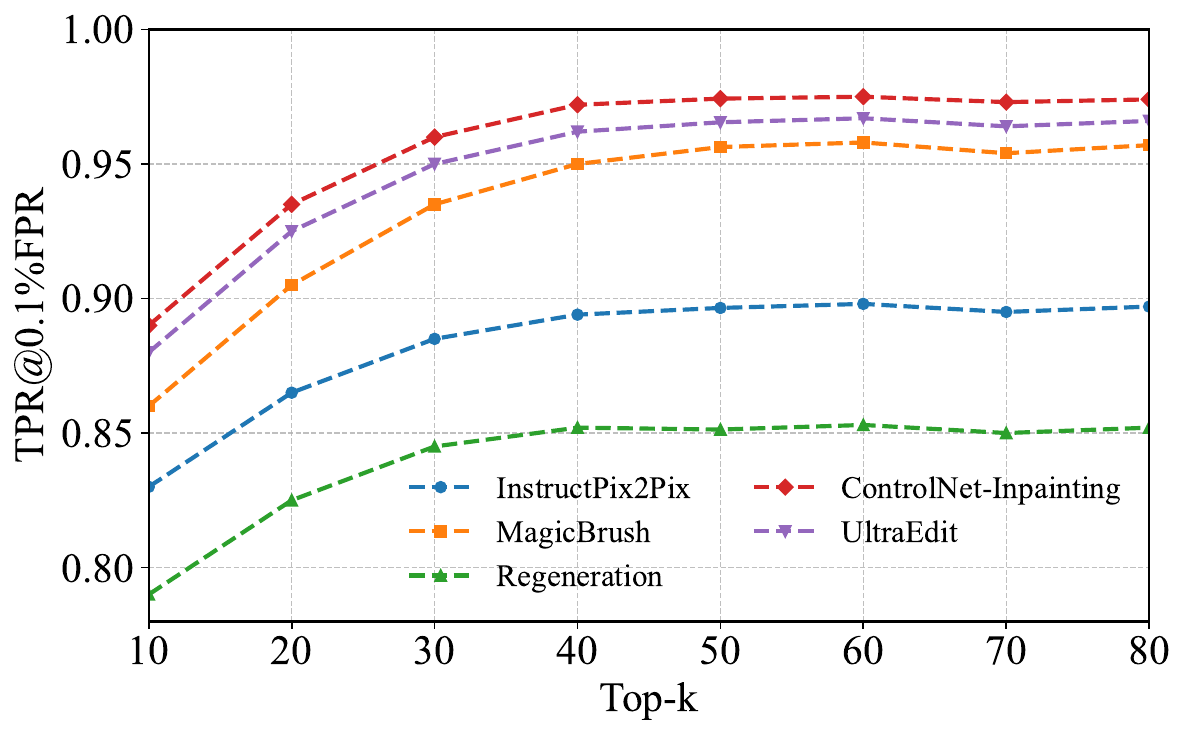}
    \caption{Impact of $K$ (Top-$K$ pairs)}
    \label{fig:topk}
  \end{subfigure}
  \hfill
  \begin{subfigure}{0.5\textwidth}
    \centering
  \includegraphics[width=0.75\linewidth]{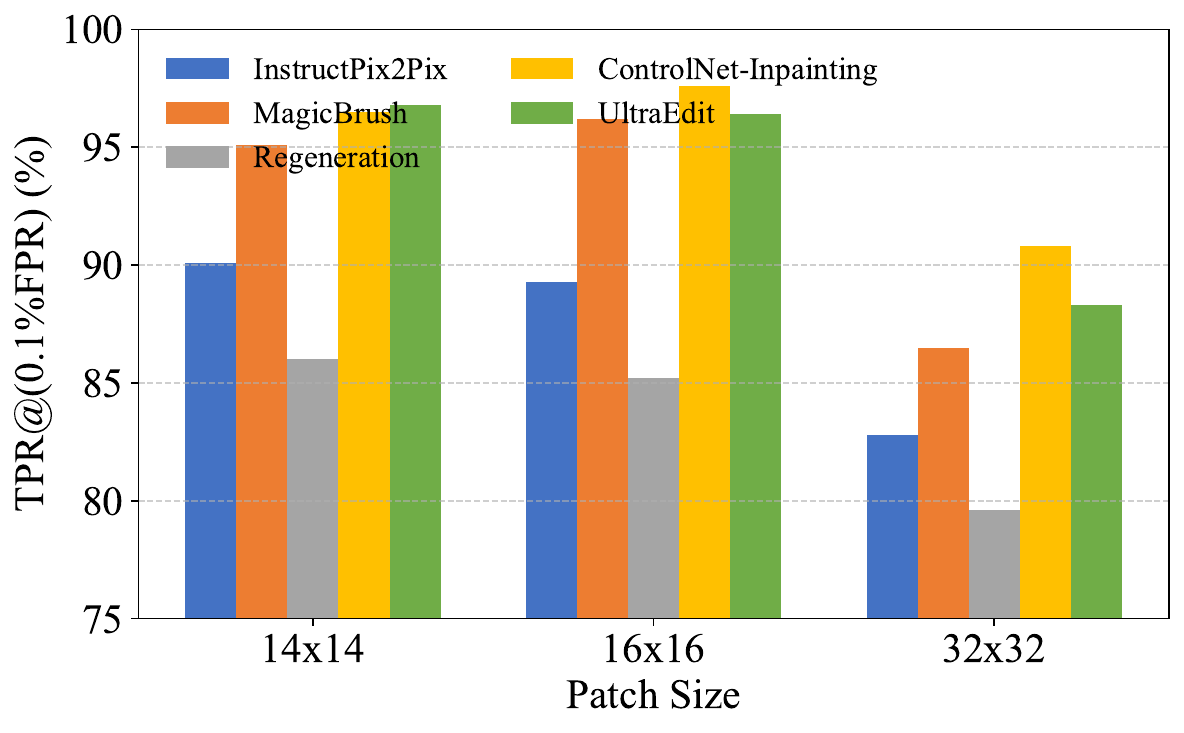}
    \caption{Impact of $N$ (Patch count)}
    \label{fig:patch}
  \end{subfigure}
  \caption{Analysis on the number of top-$K$ pairs and the patch count $N$. Robustness is evaluated over all edit types.}
  \label{fig:ablation_k_patch}
\end{figure}

Overall, \textbf{Rel-Zero} achieves strong cross-image uniqueness: 
the overlaps between zero-watermarks from different images are negligible, implying an extremely low probability of ownership collision.

\subsection{Parameter Analysis and Ablation Study }
\label{ssec:ablation}

We conduct a series of experiments to validate our design choices.

\textbf{Impact of Top-$K$.}
As shown in Fig.~\ref{fig:topk}, robustness steadily improves as the number of selected top-$K$ pairs grows, but the performance gain saturates beyond $K$ = 50. This indicates that the relational redundancy in the watermark is sufficient for stable detection, and incorporating additional pairs yields diminishing returns. Robustness is highest when it comes to ControlNet-Inpainting and UltraEdit, while Regeneration remains the most challenging due to large-scale semantic changes.

\textbf{Impact of Patch Size.}
As shown in Fig.~\ref{fig:patch}, the 14×14 and 16×16 patch sizes yield comparable robustness, as their similar granularity preserves sufficient relational detail. In contrast, performance drops sharply when the patch size increases to 32×32, indicating that overly coarse partitioning weakens the capacity to model relational structure. The reduced number of patch pairs limits relational redundancy, making the pairing too sparse for the predictor to capture fine-grained semantic invariants.

\paragraph{Component Ablation.}
To validate our architectural design, we conduct an ablation study on the framework. 
Specifically, we evaluate the robustness of different architectures under ControlNet-Inpainting. 
As shown in Table~\ref{tab:ablation_components}, replacing the ViT backbone with either ResNet-18 or ResNet-50~\cite{resnet} leads to a notable drop in robustness. 
Although ResNet variants possess strong feature extraction capability, the ViT backbone provides richer low-level priors and captures finer-grained contextual relations, which are more suitable for detecting the relative distance changes between pre- and post-edit patches. 
Consequently, ViT yields more stable and resilient relational pairs.

On the other hand, we further experiment with adding Transformer or Graph Attention Networks(GAT)~\cite{gat} layers  to the pair predictor. 
These attention-based designs turn out to provide no extra gains. 
We hypothesize that the pair prediction task primarily requires accurate estimation of patch-pair distance variations, whereas the attention mechanisms in Transformer or GAT tend to blend patch-wise representations, thereby blurring subtle relational differences and hindering precise distance discrimination.


\begin{table}[t]
  \centering
  \caption{Component ablation study. TPR@(0.1\%FPR) over ControlNet-Inpainting.}
  \label{tab:ablation_components}

  \resizebox{0.7\linewidth}{!}{
  \begin{tabular}{@{}lc@{}}
    \toprule
    Model Configuration & TPR@(0.1\%FPR) \\
    \midrule
    \textbf{-Ours (ViT + MLP)} & \textbf{97.43} \\
    \midrule
    - ViT $\rightarrow$ ResNet-18  & 84.13 \\
    - ViT $\rightarrow$ ResNet-50 & 85.21 \\
    \midrule
    - MLP $\rightarrow$ Transformer+MLP & 92.11 \\
    - MLP $\rightarrow$ GAT+MLP & 94.45 \\
    \bottomrule
  \end{tabular}
  }
\end{table}

%% file: sec/6_conclusion.tex
\section{Conclusion}

In this work, we uncover a key insight: the relational distance between partial image patches remains remarkably invariant under generative editing. We then
introduce Rel-Zero, a relational zero-watermarking framework that leverages this invariant structure to maintain robustness against diverse generative edits without compromising the image quality. Extensive experiments over multiple editing models and distortion scenarios demonstrate that Rel-Zero consistently achieves promising performance in robustness and visual fidelity. We believe this relational invariance provides a promising foundation for reliable provenance tracking in high-fidelity domains and further robust zero-watermarking paradigms in the era of large-scale generative editing models.

%% file: sec/acknowledgement.tex
\section*{Acknowledgements}
This work is supported by the National Key Research and Development Program of China under Grant 2024YFE0203200, the Strategic Priority Research Program of the Chinese Academy of Sciences(NO. XDB0690302), the National Nature Science Foundation of China under Grant U24A20329, 62527810 and 62371450.

%% file: sec/X_suppl.tex
\clearpage
\setcounter{page}{1}
\setcounter{section}{0}
\renewcommand{\thesection}{\Alph{section}}
\renewcommand\thefigure{\arabic{figure}} 
\setcounter{figure}{0}
\renewcommand\thetable{\arabic{table}}    
\setcounter{table}{0}
\maketitlesupplementary

\setcounter{equation}{0}
\renewcommand\theequation{\arabic{equation}}
\begin{figure*}[!ht]
 \centering
  \includegraphics[width=\linewidth]{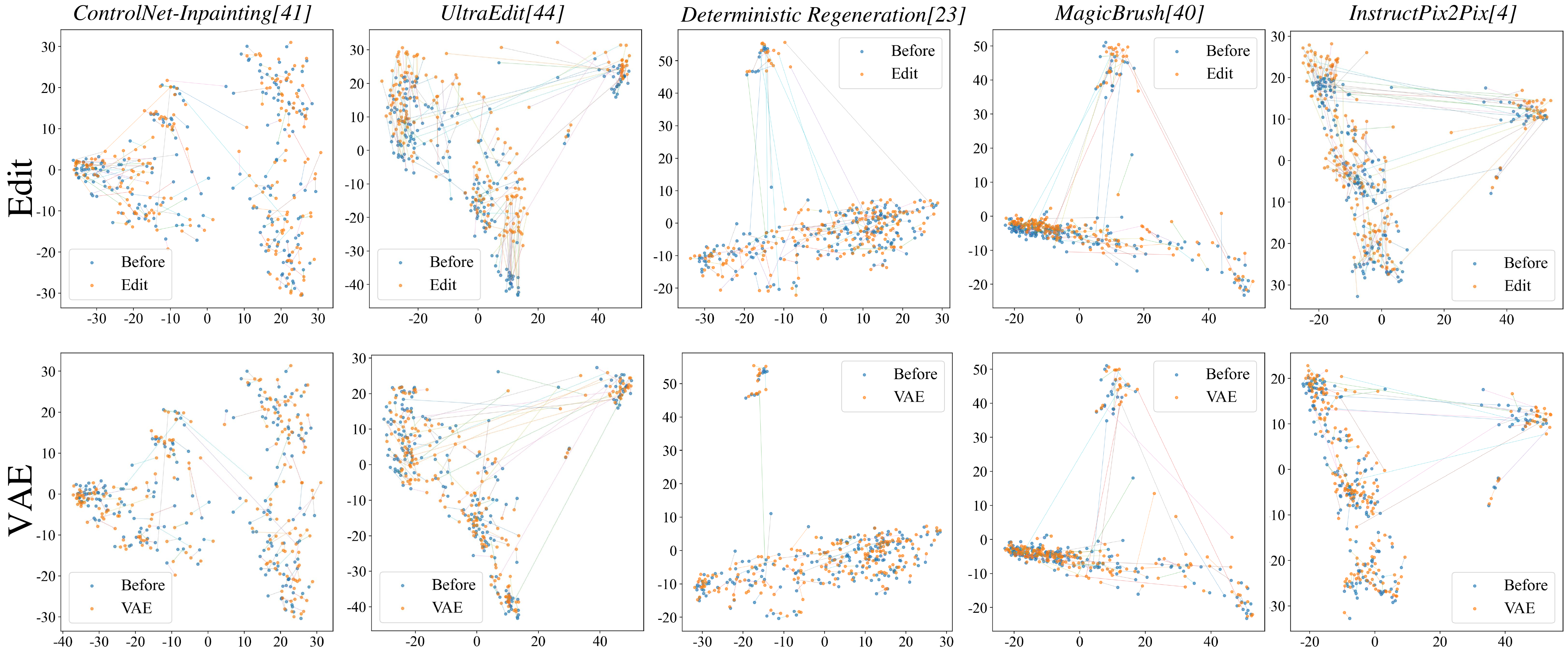}
  \caption{PCA visualization of ViT patch embeddings before and after generative editing or VAE reconstruction. Both transformations cause similar displacement patterns: only patches in drastically semantically edited regions move noticeably, while most patches remain stable. The similarity of these trajectories demonstrates that VAE reconstruction effectively mimics edit-induced feature changes.}
  \label{fig:pca}
\end{figure*}

\section{PCA Analysis Between Generative Editing and VAE}
\label{sec:vae}

\textbf{ViT Feature Visualization.}
We conduct a PCA analysis to visualize how ViT patch embeddings change after two types of transformations: (1) generative editing including ControlNet-Inpainting, UltraEdit, Deterministic Regeneration, MagicBrush, and InstructPix2Pix and (2) Stable Diffusion VAE reconstruction. For each image, we extract patch embeddings before transformation and after either editing or VAE. All embeddings are projected onto the top principal components, and we plot the movement of each patch in this low-dimensional space.

The results show a clear pattern:
the global displacement of ViT features induced by VAE reconstruction closely resembles the displacement caused by generative editing.
In both cases, only a small subset of patches—typically those corresponding to drastically semantically modified regions—exhibits noticeable shifts, while the majority of background patches remain tightly clustered and nearly unchanged. Moreover, the direction and magnitude of patch movements follow similar trajectories for Edit and VAE.

This strong resemblance indicates that VAE reconstruction preserves the structural behavior of ViT embeddings under real generative edits, despite being a much simpler and deterministic transformation. Therefore, using VAE-induced feature perturbations as a surrogate for analyzing edit-induced changes is both meaningful and well-justified.

\section{External Storage Encryption of Relational Zero-Watermark}
\label{sec:encrypt}

To protect the relational zero-watermark $\mathcal{E}_p$, which contains only
patch-pair indices but may still leak structural information, we apply a
key-controlled permutation encryption. The index set is first converted into a
binary indicator vector $\mathbf{b} \in \{0,1\}^{M}$, where $M={P \choose 2}$.
Reshaping $\mathbf{b}$ into an $N \times N$ grid $(N^2=M)$ produces a
two-dimensional watermark map suitable for keyed permutation.

We use a secret-key Arnold transform to scramble the spatial layout. Each
coordinate $(x, y)$ in the map is permuted to $(x', y')$ according to:
\begin{equation}
\begin{pmatrix}
x' \\ y'
\end{pmatrix}
=
\begin{pmatrix}
1 & p \\
q & 1
\end{pmatrix}
\begin{pmatrix}
x \\ y
\end{pmatrix}
\mod N ,
\label{eq:arnold}
\end{equation}
where $(p, q, N)$ jointly define the secret key $K$. Repeating this transform
for $T$ iterations yields the final encrypted watermark:
\begin{equation}
\mathbf{Z}_W = \text{Arnold}^{T}(\mathbf{b}; K).
\end{equation}

The encrypted $\mathbf{Z}_W$ does not reveal the patch relations or the
underlying image content, and cannot be inverted without the secret key. It can
thus be safely stored externally (e.g., hashed or registered in a protected
database) and later decrypted using the same key during verification.

\label{sec:encrypt}
\section{Time and Memory Cost}
Our method further benefits from high computational efficiency. 
Embedding-based watermarking typically requires running generative models or diffusion sampling during embedding or extraction, resulting in large runtime and memory overhead. 
Existing zero-watermarking methods also involve heavy CNN backbones or reconstruction modules.

A detailed comparison is shown in Table~\ref{tab:time_mem}. Rel-Zero avoids these costs entirely. 
All operations of patch embedding and relational comparison are implemented as a single feed-forward pass with lightweight tensor broadcasts. 
Consequently, Rel-Zero achieves a total extraction time of \textbf{0.3 ms} and a memory cost of only \textbf{0.3 GB}, outperforming prior zero-watermarking approaches by an order of magnitude and embedding-based schemes by more than two orders of magnitude.

\label{sec:time}
\begin{table}[h]
\centering
\caption{
Runtime and memory comparison. 
“Total Time” measures end-to-end watermark extraction per image (or embedding+extraction for embedding-based methods). 
CPU-only implementations (e.g., DWT-DCT and FGPCET) report no GPU memory usage and are omitted for fairness.
}
\begin{adjustbox}{max width =0.55\textwidth}
\begin{tabular}{@{}lcc@{}}
\toprule
\textbf{Method} 
& \textbf{Total Time (s)} 
& \textbf{Memory (GB)} 
\\
\midrule

DWT-DCT~\cite{dwt} 
& -- 
& -- \\

Robust-Wide~\cite{robustwide} 
& 0.0268 
& 3.1 \\

VINE~\cite{vine} 
& 0.0674 
& 5.2 \\

ConZWNet~\cite{zero1} 
& 0.0013 
& 2.3 \\

FGPCET~\cite{zero2} 
& - 
&- \\

\rowcolor{gray!20}
Rel-Zero (Ours) 
& 0.0003 
& 1.5 \\
\bottomrule
\end{tabular}
\end{adjustbox}

\label{tab:time_mem}
\end{table}

\section{Additional Robustness Experiments}

\label{sec:moreexperiment}

To further evaluate the stability of relational cues under low-level corruptions, 
we compare Rel-Zero and baseline methods under three categories of common distortions:
(1) \textbf{Salt\&Pepper noise}, applied with corruption probabilities $p{=}0.01$ and $p{=}0.03$; 
(2) \textbf{JPEG compression}, evaluated at quality factors $Q{=}90$ (mild compression) and $Q{=}50$ (strong compression);
and (3) \textbf{Rotation}, using angle perturbations of $3^\circ$ and $5^\circ$.
All distortions are applied directly to the input image without any pre-alignment or post-processing. 

Across all distortion types and strengths, Rel-Zero consistently maintains high robustness.
These results confirm that relational differences between patch pairs remain highly stable even under pixel-level perturbations, 
and that the relational design provides strong inherent robustness without requiring additional denoising, error correction, or adversarial training.

\begin{table}[t]
\centering
\caption{
Robustness under common distortions measured by TPR@(0.1\%FPR). 
We report results under Salt\&Pepper noise, JPEG compression, and small-angle rotation.
}
\resizebox{\linewidth}{!}{%
\begin{tabular}{ccccccc}
\toprule
\multirow{2}{*}{\textbf{Method}} &
\multicolumn{2}{c}{\textbf{Salt \& Pepper}} &
\multicolumn{2}{c}{\textbf{JPEG}} &
\multicolumn{2}{c}{\textbf{Rotation}} \\
\cmidrule(lr){2-3}
\cmidrule(lr){4-5}
\cmidrule(lr){6-7}
& \textbf{S\&P-1} & \textbf{S\&P-2} 
& \textbf{Q = 90} & \textbf{Q = 50}
& \textbf{Rot-$3^\circ$} & \textbf{Rot-$5^\circ$} \\
\midrule

\multicolumn{7}{l}{\textbf{Embedding Watermarking}}\\
DWT-DCT~\cite{dwt}
& 85.56 & 80.21 & 68.53 & 70.21 & 0.02 & 0.05 \\

Robust-Wide~\cite{robustwide}
& 98.89 & 98.55 & 93.55 & 95.76 & 2.43 & 3.56 \\

VINE~\cite{vine}
& 100.00 & 100.00 & 99.45 & 99.56 & 5.43 & 6.78 \\

\multicolumn{7}{l}{\textbf{Zero Watermarking}}\\
ConZWNet~\cite{zero1}
& 99.11 & 99.59 & 98.56 & 99.28 & 98.13 & 96.45 \\

FGPCET~\cite{zero2}
& 99.06 & 98.65 & 98.79 & 99.19 & 99.98 & 99.66 \\

\rowcolor{gray!20}
\textbf{Rel-Zero}
& 100.00 & 99.86 & 98.28 & 99.00 & 95.54 & 90.21 \\

\bottomrule
\end{tabular}
}
\label{tab:distort}
\end{table}

\section{ViT Embedding Stability Under Generative Editing}
\label{sec:vit_relational_stability}

The core premise of our method is that modern generative editing models 
(InstructPix2Pix, UltraEdit, etc.) tend to preserve the \emph{relational geometry}
of deep features, even when the edited image exhibits large visual or semantic changes. 
In this section, we provide additional evidence showing that the
\textbf{pairwise relationships between ViT patch features remain stable}
after editing, analogous to the relational behavior previously observed
in RGB vectors. This structural stability directly justifies our use of
patch–pair distance of ViT features as reliable watermark carriers.
\begin{figure*}[ht]
    \centering
    \includegraphics[width=\linewidth]{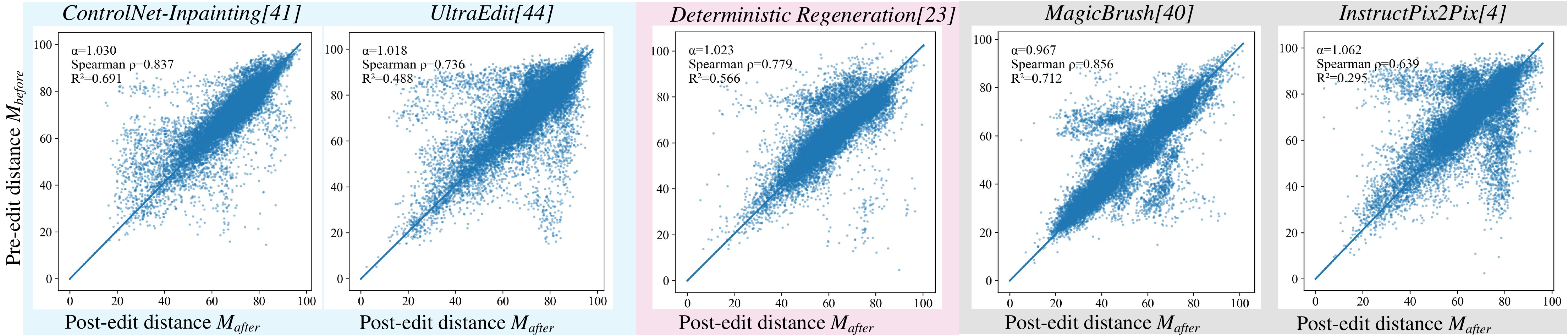}
    \caption{\textbf{Correlation between patch–pair ViT features distance before and after editing.}
    Distance aligns strongly with a fitted linear model 
    $M_{\text{after}} \approx \alpha\, M_{\text{before}}$,
    confirming that ViT patch–pair relationships remain stable and support 
    our relational watermark extraction.}
    \label{fig:vit_scatter}
\end{figure*}
\begin{figure}[ht]
    \centering
    \includegraphics[width=\linewidth]{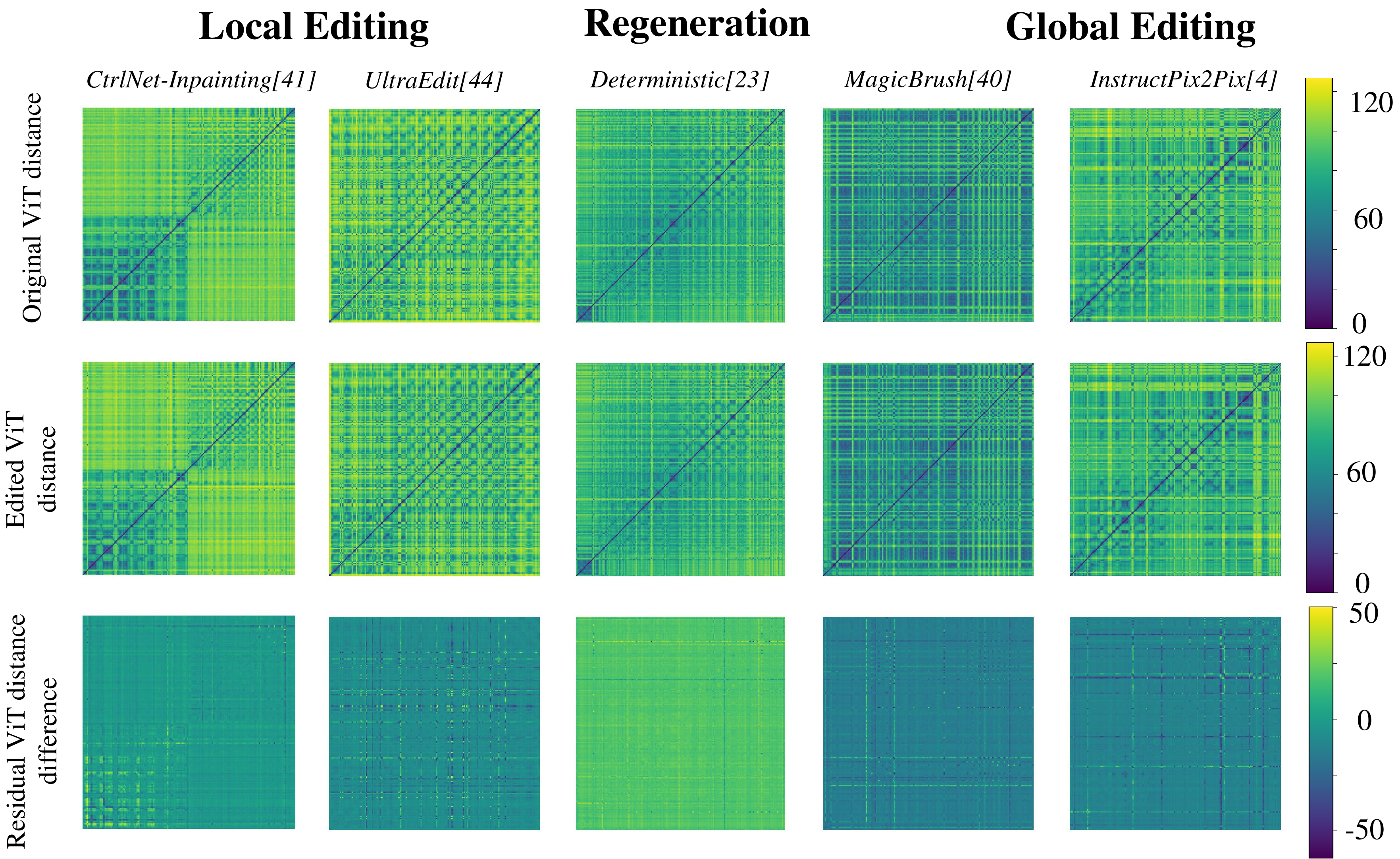}
    \caption{\textbf{ViT Self-Similarity Matrix stability under editing.}
    First row: SSM before editing. Second row: SSM after editing.
    Right: Residual $|M_{\text{after}} -  M_{\text{before}}|$.
    The near-zero residual indicates that the relational geometry of ViT 
    patch features is preserved up to a global scale.}
    \label{fig:vit_ssm}
\end{figure}
\vspace{4pt}

\noindent\textbf{Self-Similarity Matrix (SSM) Stability for ViT Features.}
Given a ViT feature map $F = \{f_i\}_{i=1}^N$ extracted from an image,
we compute the self-similarity matrix:
\[
M(i,j) = \| f_i - f_j \|_2.
\]
Figure~\ref{fig:vit_ssm} visualizes $M_{\text{before}}$ and
$M_{\text{after}}$ for the same image before and after generative editing.
Despite significant local modifications (e.g., object replacement or style
alteration), the global structure of $M$ remains nearly unchanged, up to
a global scaling factor $\alpha$ that we estimate via linear regression.
The residual matrix
\[
|M_{\text{after}} -  M_{\text{before}}|
\]
remains close to zero across most regions, demonstrating that \emph{the
intrinsic geometric layout of patch features is preserved}.  
This mirrors the behavior previously observed in pixel-space (RGB)
self-similarity patterns, but the effect is more pronounced in deep
representations due to their semantic stability.

\vspace{4pt}
\noindent\textbf{Patch–Pair Distance Correlation.}
To quantify the consistency of feature geometry, we also plot the
pairwise condensed distance $M(i,j) = \|f_i - f_j\|_2$ before and after
editing in a scatter diagram (Figure~\ref{fig:vit_scatter}).
The majority of the scatter points lie tightly along a line
\[
M_{\text{after}} \approx \alpha\, M_{\text{before}},
\]
achieving high Pearson correlation and high
coefficient of determination.  
This strong linear relationship indicates that editing operations largely
preserve the \emph{relative} differences between patch features, even if
absolute feature values shift.

\noindent\textbf{Implications for Relational Watermarking.}
Together, the SSM visualizations and distance–correlation scatterplots
demonstrate that ViT feature geometry is
\textbf{stable under a wide range of generative edits}, same as the RGB vectors in Sec~\ref{sec:theory}.
Since our watermark is based on the ordering and relationships of
patch–pair distances, and these relationships remain invariant up to
scale, the feature relational structure survives editing.  
These results provide strong empirical support for the theoretical basis
of Rel-Zero and explain why our method remains robust across global and
local edits.


\label{sec:vitanalysis}

\section{TPR with a Fixed FPR}
\label{sec:tpr}
We treat all embedding-based watermarking methods as single-bit schemes, with a embedded watermark $s \in \{0,1\}^k$. A predefined threshold $\tau \in [0,k]$, is used for detection. If the similarity score $\text{Acc}(s, s')$ between the original watermark $s$ and the extracted one $s'$ surpasses or equals $\tau$, the image is deemed as watermarked.

According to prior work~\cite{tpr}, it is generally assumed that the extracted bits $s'_1, \ldots, s'_k$ from unmarked images follow an independent and identically distributed Bernoulli process with success probability $0.5$. Under this assumption, the similarity metric $\text{Acc}(s, s')$ conforms to a binomial distribution with parameters $(k, 0.5)$.

Once this distribution is known, the false positive rate (FPR) corresponds to the likelihood that a non-watermarked image still achieves a score above the threshold $\tau$. This can be formally represented using the regularized incomplete beta function $B_x(a, b)$ as follows:
\begin{equation}
\begin{aligned}
\text{FPR}(\tau)
&= P(\text{Acc}(s, s') > \tau)
= \sum_{i = \tau+1}^{k} \binom{k}{i} \left(\frac{1}{2}\right)^k \\
&= B_{1/2}(\tau+1, k - \tau).
\end{aligned}
\end{equation}

For zero-watermarking method, we treat our relational watermarking scheme as a binary detection problem, where a fixed relational index set $\mathcal{E}_{p}$ (consisting of $K$ patch–pair edges) serves as the watermark signature. During verification, the detector extracts a candidate edge set $\mathcal{E}_{p}'$ from the query image and computes the similarity score
\begin{equation}
\eta = \frac{|\mathcal{E}_{p} \cap \mathcal{E}_{p}'|}{K}.
\end{equation}
A detection threshold $\tau \in [0,1]$ is predetermined, and the image is deemed watermarked if the overlap ratio $\eta$ reaches or exceeds this threshold.

Following common practice in watermark detection, we model the prediction of watermark edges from clean (unrelated) images as random and independent activations.
That is, each true watermark edge in $\mathcal{E}_{p}$ is falsely activated with a small probability $0.5$, and the $K$ detection outcomes are assumed to follow a binomial distribution with parameters $(K, 0.5)$.

Once this distribution is established, the false positive rate (FPR) corresponds to the probability that a clean image still produces an overlap ratio above the threshold $\tau$:
\begin{equation}
\mathrm{FPR}(\tau)
    = P(\eta \ge \tau)
    = P\!\left(X \ge \tau K\right)
    = \sum_{i=\lceil \tau K \rceil}^{K}
      \binom{K}{i} p^{\, i} (1-p)^{K-i}.
\end{equation}

In our evaluation, we maintain the FPR at $0.1\%$ to determine the corresponding operating threshold $\tau$, and subsequently report the true positive rate (TPR) computed over $1{,}000$ watermarked images. As shown in Table~\ref{tab:compare}, at a fixed FPR of $10^{-3}$ our relational watermarking method achieves strong TPR and reliable detection performance.

\section{AI-Editing Setting}
\label{sec:setting}
For deterministic regeneration, we employ the fast sampler DPM-Solver~\cite{det} and evaluate with a sampling step setting of $n_d = 25$.

We use each model’s default sampler and perform 50 sampling steps to generate edited images.
For global editing, the difficulty level is controlled by the classifier-free guidance scale of the text prompt~\cite{ho2022classifier}, which we set to 8, while fixing the image guidance scale to 1.5. 
For local editing, difficulty is determined by the ratio of the edited region to the entire image, controlled through the region mask size with intervals of 10--20\%, 20--30\%, 30--40\%, 40--50\%, and 50--60\%. 
Across all local editing settings, the image and text guidance scales are fixed at 1.5 and 7.5, respectively.

\section{Visual Comparison of Embedding-Based and Zero-Watermarking Methods}
\label{sec:quality}

\begin{figure*}[t]
    \centering
    \includegraphics[width=\linewidth]{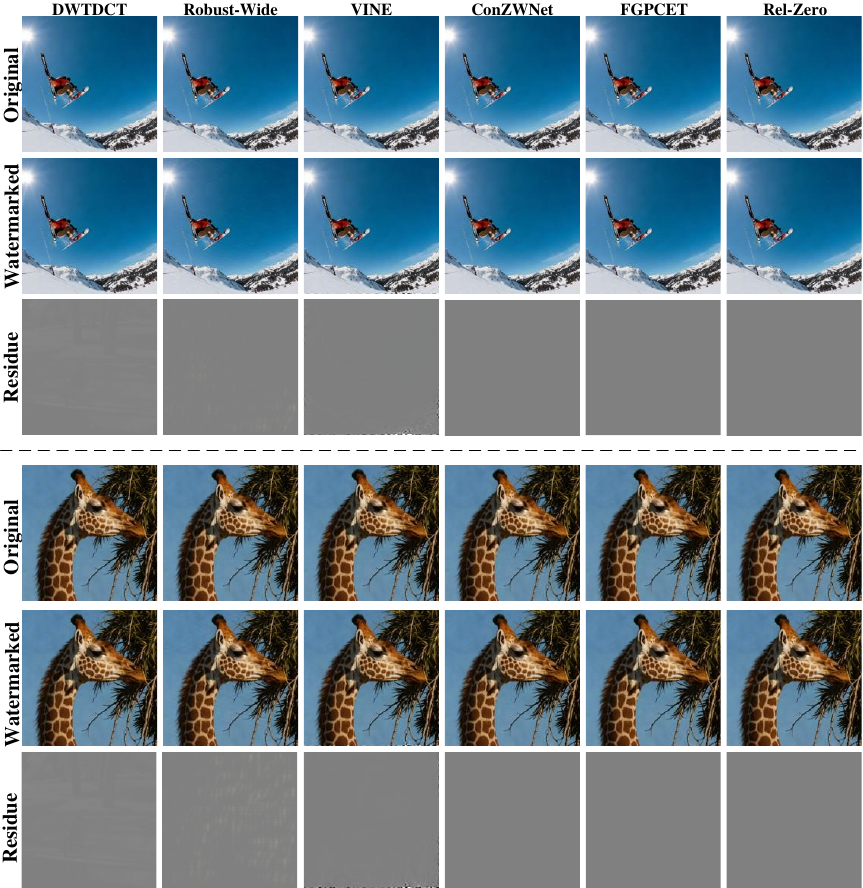}
    \caption{\textbf{Visual comparison of embedding-based watermarking and zero-watermarking.} 
    From left to right: original image, embedding-based watermarked image, residue highlighting injected perturbations, and our zero-watermarking visualization. 
    Unlike embedding-based methods, zero-watermarking introduces \emph{no pixel-level noise}, preserving perfect perceptual fidelity. Zoom in to see the detailed difference.}
    \label{fig:vis_compare}
\end{figure*}

To qualitatively assess the perceptual impact of different watermarking paradigms, we provide a visual comparison consisting of the original image, the embedding-based watermarked image, its corresponding residues, and the visualization of zero-watermarking methods (Fig.~\ref{fig:vis_compare}). 

Embedding-based watermarking inevitably introduces pixel-level perturbations to encode information. Although these perturbations may appear imperceptible in the watermarked image, they accumulate into clear artifacts in the residue map, revealing the underlying distortion injected into the visual content.

In contrast, our zero-watermarking approach requires \emph{no} modification to the input image. The extracted relational structure is purely feature-driven and leaves the pixel space entirely untouched. Consequently, it produces no observable residue, highlighting the key advantage of zero-watermarking: robust verification without introducing any noise or degradation to the perceptual quality.

